\documentclass[runningheads]{llncs}
\usepackage{graphicx}
\usepackage{xcolor}
\usepackage{amsmath}
\usepackage{amssymb}
\usepackage{ctable}
\usepackage{graphicx}
\usepackage{mathtools}
\usepackage{multirow}
\usepackage{algorithm}
\usepackage{algpseudocode}
\usepackage{adjustbox}
\usepackage{tabularx}

\begin{document}
\title{A Study of Quantisation-aware Training on Time Series Transformer Models for Resource-constrained FPGAs}
\author{Tianheng Ling \and Chao Qian \and Lukas Einhaus  \and Gregor Schiele }
\authorrunning{T. Ling et al.}
\institute{University of Duisburg-Essen, Duisburg 47057, Germany \\
\email{\{tianheng.ling, chao.qian, lukas.einhaus, gregor.schiele\}@uni-due.de}}
\titlerunning{Quantisation on Time Series Transformer}
\maketitle              
\begin{abstract} 

This study explores the quantisation-aware training (QAT) on time series Transformer models. We propose a novel adaptive quantisation scheme that dynamically selects between symmetric and asymmetric schemes during the QAT phase. Our approach demonstrates that matching the quantisation scheme to the real data distribution can reduce computational overhead while maintaining acceptable precision. Moreover, our approach is robust when applied to real-world data and mixed-precision quantisation, where most objects are quantised to 4 bits. Our findings inform model quantisation and deployment decisions while providing a foundation for advancing quantisation techniques.

\keywords{IoT \and Time Series \and Transformer \and Quantisation \and FPGA}
\end{abstract}
\section{Introduction}
\label{sec:intro}

Time series analysis is crucial in the Internet of Things (IoT), encompassing tasks such as environmental monitoring and trend prediction in smart cities. However, the complexity of real-world phenomena, amplified by factors like human activities and global climate change, poses challenges for traditional linear models to capture the non-linear and intricate patterns in sensor data. Machine learning models \cite{lara2021experimental}, particularly \emph{Transformer} models \cite{wen2022transformers}, have garnered significant attention for their effective time series modelling, efficient handling of long sequences, and ability to capture global dependencies. Despite efforts to reduce model complexity \cite{li2019enhancing}, the overhead associated with Transformers hampers their deployment on IoT devices with limited resources.

This work uses model quantisation to alleviate the overhead, particularly when deploying Transformers on low-power embedded Field-Programmable Gate Arrays (FPGAs). Model quantisation involves mapping computations from high-resolution floating-point numbers (e.g., 32-bit) to lower-resolution integer/fixed-point numbers (e.g., 8-bit, 4-bit, or 2-bit) through the application of quantisation schemes. Previous research \cite{wojcicki2022accelerating} used post-training quantisation (PTQ) to select the optimal resolution per layer, resulting in mixed-precision quantisation. In contrast, this work focuses on the selection of quantisation schemes. Specifically, we assess the effects of symmetric and asymmetric quantisation schemes on Transformer models and suggest enhanced quantisation-aware training (QAT) to select the optimal quantisation scheme for each object dynamically. We evaluate this approach with mixed-precision quantisation and examine the trade-off between the computational overhead reduction and the prediction precision loss.

The subsequent sections are structured as follows: Section 2 discusses related work, followed by background information on time series Transformer models in Section 3. Section 4 presents considerations for quantising the linear layers in the Transformer model. Section 5 elaborates on our approach for conducting this study. The experimental setup and results are provided in Section 6. Finally, Section 7 concludes the paper, summarises our findings, and outlines future research.

\section{Related Work}

While the quantisation of Transformer models in Natural Language Processing has been extensively studied \cite{qin2022bibert}, its application to time series analysis has received limited attention. The unique challenges posed by differences in input/output representation, data processing, and target definition between text and time series necessitate tailored quantisation techniques. Therefore, existing quantisation methods developed for text-based models cannot be directly applied to time series analysis. 

Only a little research focuses on quantising time series Transformer models. One notable study by Becnel et al. \cite{becnel2022tiny} investigated the quantisation of a Transformer model called T$^3$ for predicting environmental data in univariate and multivariate settings. They successfully reduced the model size from 16-bit floating-point to 8-bit using PTQ with TensorFlow Lite. When deploying the quantised model on an ESP32 microcontroller, they could compress it to 68 KB. However, the root mean square error (RMSE) increased, ranging from 23.824\% to 92.632\%, depending on the target variable. In another study, Wojcicki et al. \cite{wojcicki2022accelerating} focused on mixed-precision quantisation through PTQ to strike a balance between inference speed and accuracy in a Transformer model for hadronic jet tagging classification. Their findings demonstrated significant acceleration of the Transformer model on XCU250 FPGAs compared to GPUs. However, it is worth noting that most layers in their approach required more than 6-bit quantisation to achieve optimal model precision.

Although previous studies have touched upon the quantisation of Transformer models for time series analysis, they have yet to fully incorporate QAT and explore lower-bit quantisation, such as 4 bits or less. We aim to address this gap by recognising the significance of these aspects in facilitating the deployment of Transformer models on embedded FPGAs with limited resources. 

\section{Time Series Transformer}
\label{sec:tranformer}
\begin{figure}[!htb]
    \centering
    \includegraphics[width=0.68\textwidth]{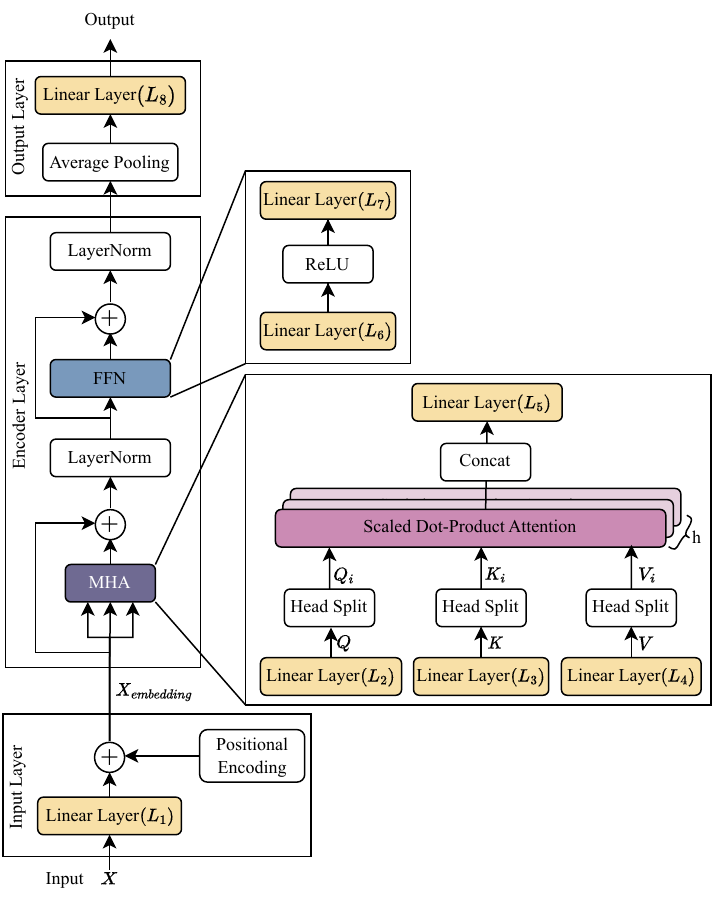}
    \caption{The Architecture of the Transformer Model For Time Series Prediction}
    \label{fig:transformer}
\end{figure}

This section presents a Transformer model explicitly designed for single-step ahead time series forecasting. We use this model in our study. As shown in Figure \ref{fig:transformer}, the architecture of the Transformer model consists of an input layer, an encoder layer, and an output layer. This model is based on the T$^{3}$ model introduced in \cite{becnel2022tiny}. However, we modified their original multi-target model to create a single-target model, enabling us to focus on the quantisation aspect. It is essential to highlight that extending our research to incorporate multi-target prediction can be easily pursued in future work.

We assume input $X$ with dimensions $(n,m)$ represents a sequence of $n$ continuous data points with $m$ dimensions, where $m\geq 1$ accommodates univariate and multivariate time series. The input layer applies the transformation of $X$ through a linear layer (labelled $L_{1}$). Then it incorporates positional information using the Positional Encoding (PE) operation to produce $X_{embedding}$ with dimensions $(n,d_{model})$. Subsequently, $X_{embedding}$ is fed into the encoder layer for further processing. This encoder layer consists of a Multi-head Self-attention (MHA) module and a Feedforward Network (FFN) module, followed by a Skip Connection and Layer Normalisation (LN). Within the MHA module, three linear layers (labelled $L_{2}$, $L_{3}$ and $L_{4}$ respectively) independently transform $X_{embedding}$ into the Query ($Q$ with dimensions $(n,d_{Q})$), Key ($K$ with dimensions $(n,d_{K})$), and Value ($V$ with dimensions $(n,d_{V})$) matrices. The three matrices are subsequently fed into the computation of Scaled Dot-Product Attention. The resulting outputs are then processed through a linear layer (labelled $L_{5}$) to obtain the output (with dimensions $(n,d_{O})$)) of the MHA module. The FFN module comprises two linear layers, denoted as $L_{6}$ and $L_{7}$, whose outputs with dimensions $(n, d_{ffn})$ and $(n, d_{model})$ respectively. The activation function applied between these layers is ReLU. Lastly, the output layer uses average pooling and a linear layer (labelled $L_{8}$) to generate an output $y$. 

\begin{table}[!htb]
\renewcommand\arraystretch{1.2}
\tabcolsep=0.1cm
\centering
\caption{Trainable Parameters of the Transformer Model}
\begin{adjustbox}{center}
\begin{tabular}{cccc}
\specialrule{.2em}{.1em}{.1em}
\multicolumn{2}{c}{\textbf{Module}} & \textbf{Layer/Operation} & \textbf{Parameters} \\
\hline
\multicolumn{2}{l}{Input Layer} & Linear Layer ($L_{1}$) & $m \cdot d_{model} + d_{model}$\\
\hline
\multirow{8}{*}{\begin{tabular}[c]{@{}c@{}} Encoder \\ Layer{}\end{tabular}} &\multirow{5}{*}{MHA} & Linear Layer ($L_{2}$) & $d_{model} \cdot d_Q + d_Q$   \\
&  & Linear Layer ($L_{3}$)  & $d_{model} \cdot  d_K + d_K$  \\
&  & Linear Layer ($L_{4}$) & $d_{model} \cdot  d_V + d_V$ \\
&  & Linear Layer ($L_{5}$)  & $d_{model} \cdot  d_{O} + d_{O}$  \\
&  & LN  & $2  \cdot  n  \cdot  d_{model}$ \\ \cline{2-4}
& \multirow{3}{*}{FFN} & Linear Layer ($L_{6}$) & $d_{model}  \cdot  d_{ffn} + d_{ffn}$  \\
 & & Linear Layer ($L_{7}$)  &  $ d_{ffn}   \cdot  d_{model}+ d_{model}$ \\
 & & LN & $2   \cdot  n  \cdot  d_{model}$ \\ \hline
\multicolumn{2}{l}{Output Layer}  & Linear Layer ($L_{8}$)  & $ d_{model} + 1$   \\
\specialrule{.2em}{.1em}{.1em}
\end{tabular}
\end{adjustbox}
\label{tab:cal_params}
\end{table}

Table \ref{tab:cal_params} shows the distribution of all trainable model parameters by providing a breakdown of the trainable parameters in the linear layers and LN operations. For simplicity, we assume that $d_Q$, $d_K$, $d_V$ and $d_O$ are set to $d_{model}$, while $d_{ffn}$ is set to $4 \cdot d_{model}$. With these settings, we can calculate the total number of parameters in the Transformer model (see Equation \ref{eq:total_param}) and the total number of parameters in all linear layers (see Equation \ref{eq:linear_param}).

\begin{align}
\text{Params}_\text{Total} & = 1 + (11+m+4  \cdot  n)  \cdot  d_{model} + 12  \cdot d^2_{model} \label{eq:total_param} \\
\text{Params}_\text{Linear}  & = 1 + (11+m)  \cdot  d_{model} + 12   \cdot  d^2_{model}  \label{eq:linear_param} 
\end{align}

In this study, we adopt the hyperparameters used in the T$^{3}$ model. To be specific, the embedding dimension is set to $d_{model}=64$, $d_Q$, $d_K$ and $d_V$ are set to $d_{model}$, and the FFN dimension is set to $d_{ffn}=256$. Letting the model utilise seven feature variables from the preceding 24 observations to predict the target variable for the subsequent observation, the model consists of 56,449 parameters, with 50,305 parameters residing in the linear layers, accounting for 89.116\% of the total model parameters. When we only quantise the linear layers from 32 bits (floating-point) to 8 bits, the model size can be compressed by $3.015\times$. Thus, we decided to quantise the linear layers exclusively. 


\section{Quantisation of Linear Layers}
\label{sec:quant_method}

Our study focuses on affine quantisation \cite{krishnamoorthi2018quantizing,nagel2021white}, which involves mapping continuous floating-point $r\in [\beta, \alpha]$ to signed integers $q$ using $b$ bits. This mapping process can be implemented in two schemes: 1) asymmetric quantisation (AQ) and 2) symmetric quantisation (SQ), depending on whether the upper bound ($\alpha$) and lower bound ($\beta$) of the floating-point numbers are symmetric around 0.

The AQ scheme is described by Equation \ref{eq_linear_quant}, where $s$ (generated by Equation \ref{eq_asymmetric_signed_scale}) represents the scale factor between $r$ and $q$, and $z$ (calculated by Equation \ref{eq_asymmetric_signed_zero_point}) denotes the zero point, which is an integer representation of the floating-point zero. It is important to note that the zero point is not necessarily equal to integer zero under this scheme. The $\text{clip}$ function limits the result to the integer ranges of $q$ to mitigate the risk of overflow, while the $\text{round}$ function performs the nearest rounding operation. The dequantisation process of the AQ scheme is specified by Equation \ref{eq_asymmetric_DQuantise}, approximating a floating-point $r'$ from the integer $q$.

\begin{align}
q & = \text{round}(\text{clip}(\frac{r}{s}+ z; \, -2^{b-1},2^{b-1}-1)) \label{eq_linear_quant}\\
s & = \frac{\alpha - \beta}{2^b-1} \label{eq_asymmetric_signed_scale}\\
z & = \text{round}(\text{clip}(2^{b-1}-1-\frac{\alpha}{s}; \, -2^{b-1},2^{b-1}-1))\label{eq_asymmetric_signed_zero_point}\\
r & \approx r' = s(q-z) \label{eq_asymmetric_DQuantise}
\end{align} 

The SQ scheme, as a specific AQ scheme, is designed to map approximately 0-symmetric floating-point numbers to absolute 0-symmetric integers. The quantisation process is described by Equations \ref{eq_symmetric_signed_quant},  where the zero point is excluded since it equals the integer zero. To ensure absolute 0-symmetry, the smallest integer $-2^{b-1}$ is also omitted. The scale factor is determined using Equation \ref{eq_symmetric_scale}, where the range of floating-point numbers is twice the maximum absolute value between $\alpha$ and $\beta$, and the number of representable integers is one less than that in the AQ scheme. The dequantisation process is defined by Equation \ref{eq_symmetric_DQuantise}.

\begin{align}
q & = \text{round}(\text{clip}(\frac{r}{s}; \, -2^{b-1}+1, 2^{b-1}-1)) \label{eq_symmetric_signed_quant}\\
s & = \frac{2 \cdot max(|\alpha|, |\beta|)}{2^{b}-2} \label{eq_symmetric_scale}\\
r & \approx r' = sq \label{eq_symmetric_DQuantise}
\end{align}

Compared to the AQ scheme, the SQ scheme offers the advantage of eliminating computational overhead caused by zero points. However, in the SQ scheme, the smallest integer is discarded to preserve the zero symmetry of the integers, resulting in a loss of precision. This loss becomes particularly significant in the case of low-bit quantisation. For 4-bit quantisation, the absence of one of the 16 representable integers introduces a substantial decrease in precision. Therefore, when selecting quantisation schemes, it is vital to consider the trade-off between precision and computational overhead.

While quantising a linear layer, weights ($W_{float}$), biases ($B_{float}$), inputs ($I_{float}$), and outputs ($O_{float}$) could be chosen as quantisation objects. Typically, weights and biases (hereafter collectively referred to as layer parameters) in the linear layer exhibit a zero-symmetric pattern, while inputs and outputs (hereafter layer feature vectors) follow a zero-asymmetric pattern. Consequently, many relevant studies \cite{jacob2018quant,zafrir2019q8bert} utilise the SQ scheme for layer parameters and the AQ scheme for layer feature vectors. However, this assumption may only hold for some model architectures and applications, as the data distribution of quantisation objects can vary. Hence, choosing a quantisation scheme based on the actual data distribution is crucial. However, determining the data distribution becomes complex due to changes during the training process. Moreover, fixing the quantisation scheme directly before training and further searching for the best combination of schemes based on the model performance after training is highly inefficient. 

Our research aims to explore the feasibility of quantising the Transformer model by selecting the optimal quantisation scheme for each quantisation object in the linear layers. This investigation seeks to find a balance between precision and computational overhead.

\section{Our Approach}
\label{sec:implementation}
In this section, we discuss the computational overhead estimation methodology, introduce an adaptive quantisation scheme, and describe the implementation details of a custom QLinear module, including its supported meta-parameters.

\subsection{Computational Overhead Estimation}

To assess the computational overhead introduced by the AQ scheme compared to the SQ scheme during quantised inference in the Transformer model, it is important to analyse and quantify this overhead. We focus on the linear transformation process, quantising all quantisation objects adopting the AQ scheme. Equation \ref{eq:quant_linears} illustrates this transformation, where $W_{int}$ and $B_{int}$ represent the quantised layer parameters obtained during QAT and stored in the compressed model. Additionally, $I_{int}$ and $O_{int}$ denote the quantised layer feature vectors. In this context, we introduce the variables $s_W$, $s_B$, $s_I$, and $s_O$ to indicate the scale factors for weights, biases, inputs, and outputs, respectively. Similarly, the variables $z_W$, $z_B$, $z_I$, and $z_O$ represent the corresponding zero points. The computations involving $(W_{int}-z_W)$ and $(B_{int}-z_B)$ can be pre-computed prior to inference, resulting in no computational overhead. However, the computations of $(I_{int}-z_I)$ and $(O_{int}-z_O)$ are unavoidable during inference. The magnitude of this overhead is directly influenced by the dimensions of $I_{int}$ and $O_{int}$. 

\begin{align}
s_O(O_{int} -z_O) & = s_I(I_{int}-z_I)s_W(W_{int}-z_W) + s_B(B_{int}-z_B) \label{eq:quant_linears} 
\end{align}

Taking the linear layer ($L_{1}$) of the model input layer as an example, with inputs $I_{int}$ with dimensions of $(n,m)$ and outputs $O_{int}$ with dimensions of $(n,d_{model})$, the AQ scheme applied to both the inputs and outputs introduce a computational overhead of $n \cdot m + n \cdot d_{model}$ operations. Alternatively, if either the inputs or the outputs undergo the AQ scheme, the computational overhead is $n \cdot m$ or $n \cdot d_{model}$ operations, respectively. The same principle is applicable to the computational overheads of the other linear layers. 

In the subsequent sections, we will assess the overall computational overhead by summing the overheads of all linear layers based on their respective quantisation scheme combinations. This estimation method allows us to evaluate the effectiveness of different quantisation scheme combinations.

\subsection{Adaptive Quantisation Scheme}
\label{subsec:adaptiv}

To explore the impact of different combinations of quantisation schemes (referred to as quantisation configuration) on the model performance, it is critical to have a comprehensive coverage of samples representing these configurations. For this study, we manually selected three quantisation configurations presented in the first three rows of Table \ref{table:diff_exp}. Notably, the `All-AQ' configuration applies the AQ scheme to all layer parameters and feature vectors, resulting in the highest computational overhead. Conversely, the `All-SQ' configuration aims to minimise computational overhead but may suffer from precision loss. Additionally, the `SQ+AQ' configuration is designed based on the typical data distribution of layer parameters and feature vectors.

\begin{table}[!htb]
\caption{Description of Different Quantisation Configurations}
\centering
\begin{adjustbox}{center}
\begin{tabular}{@{}l@{\hspace{12pt}}l@{}}
\specialrule{.2em}{.1em}{.1em}
\textbf{\begin{tabular}[c]{@{}l@{}}Quantisation \\Configuration\end{tabular}}  & \textbf{Description} \\ \hline
All-AQ &  AQ scheme applied to all parameters and feature vectors    \\
All-SQ & SQ scheme applied to all parameters and feature vectors     \\
SQ+AQ & SQ scheme for parameters, AQ scheme for feature vectors      \\
SQ+APQ & SQ scheme for parameters, APQ scheme for feature vectors    \\
\specialrule{.2em}{.1em}{.1em}
\end{tabular}
\end{adjustbox}
\label{table:diff_exp}
\end{table}

However, these three baselines may not adequately cover all quantisation configurations. As mentioned in Chapter 3, our model comprises 8 linear layers, meaning these baselines represent only 3 out of the $2^{16}$ possible combinations for our Transformer model. Therefore, we incorporate quantisation scheme awareness into the training process. This approach allows us to update the quantisation scheme, similar to the standard QAT procedure for updating quantisation parameters. We refer to this approach as the Adaptive Quantisation (APQ) Scheme. During training, after processing each mini-batch, the quantisation scheme selection is dynamically updated for individual quantisation objects as needed. Consequently, it has the potential to find the most suitable combination of quantisation schemes during QAT.

\begin{equation}
\text{quant\_scheme}=\left\{
\begin{aligned}
SQ &\text{, if }\beta > 0, \alpha < 0,\left| \frac{\beta + \alpha}{\max(\beta, \alpha)} \right| < \text{threshold} \\
AQ &\text{, if others} 
\end{aligned}
\right.
\label{eq:adaptive}
\end{equation}

We implement the selection logic of the APQ scheme by evaluating whether the distribution of floating-point numbers exhibits 0 symmetry, as defined in Equation \ref{eq:adaptive}. To control the decision-making process, we introduce a threshold value that determines the strictness in determining symmetry. This threshold value can be tailored based on the model architecture and specific application requirements, allowing for customisation. Based on the `SQ+AQ' configuration, we apply the APQ scheme to quantise the layer feature vectors, resulting in the `SQ+APQ' configuration (see row 4 in Table \ref{table:diff_exp}). By examining the performance of the resulting models for these four configurations, we aim to provide insights into quantising the Transformer model by combining different quantisation schemes.

\subsection{QLinear Implementation}

As the PyTorch framework currently supports quantisation only down to 8 bits, we have addressed this limitation by implementing a custom QLinear module. This module is a subclass of the \textit{torch.nn.Linear} module, inheriting its core functionality while expanding the range of available bit widths for data representation. Our custom module also introduces additional options for quantisation, specifically tailored for this study.

\begin{table}[!htb]
\caption{Meta-Parameters of the QLinear Module}
\centering
\begin{adjustbox}{center}
\begin{tabular}{@{}l@{\hspace{12pt}}l@{\hspace{12pt}}l@{}}
\specialrule{.2em}{.1em}{.1em}
\textbf{Meta-Parameter} & \textbf{Data Type} & \textbf{Options} \\ \hline
quantisation\_object & \textit{string} &   [weights, biases, inputs, outputs] \\
quantisation\_bits & \textit{integer} &  [2, 4, 8, 16] \\
quantisation\_scheme & \textit{string} &  [SQ, AQ, APQ] \\  
\specialrule{.2em}{.1em}{.1em}
\end{tabular}
\end{adjustbox}
\label{table:qlinea}
\end{table}

Table \ref{table:qlinea} provides a detailed overview of the quantisation objects: weights, biases, inputs, and outputs, along with their corresponding quantisation options. Users are empowered to individually assign different quantisation bit widths, ranging from 2 to 16, allowing for a diverse range of precision levels in the quantisation process. In addition to the flexibility in selecting quantisation bit widths, users can also choose the APQ scheme for specific quantisation objects. With APQ, the scheme is automatically determined during QAT. Alternatively, users can fix the quantisation scheme as either SQ or AQ before the training process. This feature enables fine-tuning and customisation according to individual requirements and preferences.

\section{Experiments and Results}
\label{sec:exp}
In this section, we introduce the dataset and experimental settings. We then present results for two study phases: evaluating model precision and computational overhead with 8-bit quantisation, and extending the analysis to mixed-precision quantisation.
\subsection{Experiments Settings}
\label{sec:exp_setting}

We used the \textit{AirU} dataset for air pollution forecasting, which was introduced in \cite{becnel2022tiny}. This publicly available dataset consists of 19,380 observations, including timestamp information and seven feature variables (PM$_{1}$, PM$_{2.5}$, PM$_{10}$, Temperature, Humidity, RED, and NOX), as well as the target variable Ozone. Our work leverages the feature variables from the preceding 24 observations to predict the target variable for the next observation. After removing discontinuous observations, we obtained 15,258 pairs of feature-target samples. To ensure a fair comparison with the work \cite{becnel2022tiny}, we selected the observations from the same period as the test set, resulting in a training set of 14,427 samples and a test set of 831 samples. Unlike previous work \cite{becnel2022tiny}, we compute the normalisation parameters using the training data to avoid any data leakage. We normalise all the data using the MinMax method.

The details of the model parameterisation can be found in Section \ref{sec:tranformer}. For model training, we conducted 100 epochs using early stopping, with a batch size of 256. We used the \textit{Adam} optimizer with parameters $\beta_1=0.9$, $\beta_2=0.98$, and $\epsilon=10^{-9}$. The initial learning rate was set to 0.01. Dropout with a rate of 0.2 was applied after the PE operation, MHA, and FFN modules. The Mean-Squared Loss function was used for training. After applying the inverse transformation to the outputs and normalised target values, we computed the RMSE on the test data to evaluate the model. The trained full-precision Transformer (FTransformer) model achieved an RMSE of 3.989, demonstrating comparable performance to the reported RMSE of 4.120 in \cite{becnel2022tiny}.

\subsection{The Effect of Quantisation Configurations on Transformer}

Experiments were conducted to evaluate the impact of different quantisation configurations (as described in Table \ref{table:diff_exp}) on model precision and computational overhead. We simplified the process in these experiments by applying 8-bit quantisation to all layer parameters and feature vectors. The results in Table \ref{tab:diff_quant_configu} demonstrate a slight increase in the RMSE of the quantised models compared to the FTransformer model. In addition, the quantised models achieved a significantly smaller model size of 73.13 KB compared to the original model size of 220.50 KB.

\begin{table}[!htb]
\renewcommand\arraystretch{1.2}
\tabcolsep=0.1cm
\centering
\caption{Performance Comparison with Different Quantisation Configurations}
\begin{adjustbox}{center}
\begin{tabular}{ccc}
\specialrule{.2em}{.1em}{.1em}
\textbf{Quantisation Configuration} & \textbf{RMSE} & \textbf{Operation Overhead} \\ \hline
\textbf{All-AQ}   &   4.009 ($\uparrow$ 0.501\%) & 29417\\
\textbf{All-SQ}   &   4.120 ($\uparrow$ 3.284\%) & 0 \\ 
\textbf{SQ+AQ}    &   4.079 ($\uparrow$ 2.256\%) & 29417\\
\textbf{SQ+APQ (1)}   &   \textbf{3.977 ($\downarrow$ 0.300\%)} & 20201 \\
\textbf{SQ+APQ (2)}  &   4.123 ($\uparrow$ 3.359\%) & \textbf{15529} \\ 
\specialrule{.2em}{.1em}{.1em}
\end{tabular}
\end{adjustbox}
\label{tab:diff_quant_configu}
\end{table}

Specifically, the `All-AQ' configuration resulted in a quantised model with an RMSE of 4.009, just 0.501\% higher than the FTransformer model. Conversely, the `All-SQ' configuration exhibited a 2.769\% higher RMSE than the `All-AQ' configuration. This discrepancy can be attributed to the introduction of error when applying the SQ scheme to quantisation objects that do not exhibit 0 symmetry. Moreover, under the `SQ+AQ' configuration, the RMSE of the quantised model was worse than the `All-AQ' but better than the `All-SQ'. This difference in RMSE across the three quantisation configurations can be attributed to the number of layer feature vectors that have chosen the AQ scheme, aligning with our expectations. In addition, both `All-AQ' and `SQ+AQ' configurations introduced the same amount of computational overhead, totalling 29,417 operations, in contrast to the `All-SQ' configuration.

Applying the `SQ+APQ' configuration incorporating a threshold of 0.1, we got two extreme cases with 100 model training, namely `SQ+APQ (1)' and `SQ+APQ (2)'. 
The `SQ+APQ (1)' case has the best model precision, achieving an RMSE of 3.977, even 0.300\% lower than the FTransformer model. Notably, it also exhibits a significant reduction of 31.329\% in computational overhead compared to the `All-AQ' configuration. Moreover, for scenarios where computational overhead is vital, the `SQ+APQ (2)' showcased its effectiveness by offering a substantial reduction of 47.211\% in computational overhead compared to the `All-AQ' configuration, albeit with a slightly inferior RMSE of 4.123 (0.073\% higher). This slight performance decline in the `SQ+APQ (2)' can be attributed to the increased complexity introduced by the adaptive approach during the training process. These findings validate the potential of our proposed approach and provide insights into the impact of different quantisation configurations on the precision and operation overhead of the Transformer model.

\subsection{Extension to Mixed-precision Quantisation}

Our next goal is to evaluate the impact of different quantisation configurations in lower-bit quantisation scenarios in pursuit of smaller model size, improving their deployment possibilities on embedded FPGAs. We begin with the `All-AQ' configuration to quantise all linear layers to 4 bits. While the compressed model size is significantly reduced by a factor of $4.541\times$, it also results in a substantial 94.660\% increase in RMSE, indicating a loss of predictive power in the quantised model.

To understand the sensitivity of each linear layer to 4-bit quantisation, we conducted ablation studies to quantise one linear layer to 4 bits while quantising the remaining layers at 8 bits. The linear layer ($L_{8}$) in the model output layer exhibits the highest sensitivity to 4-bit quantisation, contributing significantly to the degradation in RMSE when all linear layers were quantised to 4 bits. Upon analysing the output and the target values, it is observed that quantising this linear layer to 4 bits severely restricted the model's capacity to represent values of the target variable, which are discrete values ranging from 0 to 90. Hence, we opted to quantise the linear layer ($L_{8}$) using 8 bits and the remaining layers using 4 bits. This decision yielded a more acceptable RMSE increase of 15.593\% (see Table \ref{tab:extend} row 1). Importantly, the model compression rate of $4.358\times$ experienced a marginal decrease of 0.003, which is negligible given the output layer's parameter count of 65. As a result, the model size is compressed to 50.596 KB.

\begin{table}[!htb]
\renewcommand\arraystretch{1.2}
\tabcolsep=0.1cm
\centering
\caption{Performance Comparison of Mixed-Precision Quantised Linear Layers}
\begin{tabular}{ccc}
\specialrule{.2em}{.1em}{.1em}
\textbf{Quantisation Configuration} & \textbf{RMSE} & \textbf{Operation Overhead} \\ \hline
\textbf{All-AQ}     &   $4.611(\uparrow15.593\%)$   & 29417\\
\textbf{All-SQ}     &   $5.141(\uparrow28.879\%)$   & 0 \\
\textbf{SQ+AQ}      &   $4.721(\uparrow18.350\%)$   & 29417 \\
\textbf{SQ+APQ (1) } &   \textbf{4.872($\uparrow$22.135\%)}  & 26345 \\
\textbf{SQ+APQ (2) }  &   $5.049(\uparrow26.573\%)$   & \textbf{10921} \\
\specialrule{.2em}{.1em}{.1em}
\end{tabular}
\label{tab:extend}
\end{table}

We then expanded these findings of mixed-precision quantisation to other quantisation configurations. Table \ref{tab:extend} shows that the `All-SQ' configuration resulted in a quantised model with an RMSE of 5.141, 28.879\% higher than the FTransformer model. This can be attributed primarily to the fact that when quantising with 4 bits, the SQ scheme has to discard the integer -8. As anticipated, the `SQ+AQ' configuration yields the quantised model with an RMSE value between the other two configurations. The computational overhead remains unchanged from the previous experiments.

Compared to the above configurations, we highlight the capability of the `SQ+APQ' configuration by showing two models (1) aiming for higher precision and (2) aiming for lower overhead. Model (1) achieves an RMSE of 4.872, which is 5.660\% higher than the 'All-AQ' configuration. However, it offers the advantage of a reduced computational overhead of 10.443\%. Model (2) achieves a substantial reduction in overhead by 62.875\%, while its RMSE is 1.789\% lower compared to the ‘All-SQ' configuration. In summary, under mixed-precision quantisation, we further confirmed the above-obtained insights.

\section{Conclusion and Future Work}

This paper studies the impact of different quantisation schemes on the linear layers of the Transformer model for time series forecasting. We propose an adaptive method that dynamically adjusts the quantisation scheme during training, selecting the most suitable scheme for each quantisation object. Our approach is evaluated on real-world data using the Transformer model for time series prediction tasks. Our method effectively quantises the Transformer model through both pure 8-bit quantisation and mixed-precision quantisation, achieving a balance between reduced computational overhead and improved precision. In future work, we will delve deeper into the adaptive method, exploring more sophisticated approaches that employ more explainable parameters instead of intuitive thresholds.

\noindent{\textbf{Acknowledgements.}} The authors acknowledge the financial support from the Federal Ministry of Economic Affairs and Climate Protection of Germany (RIWWER project, 01MD22007C).

\bibliographystyle{splncs04}
\bibliography{reference}
\end{document}